\documentclass[runningheads]{llncs}

\usepackage{ulem}
\usepackage[square,numbers]{natbib}
\usepackage[shortlabels]{enumitem}
\setlist{leftmargin=4mm}
\usepackage[T1]{fontenc}

\usepackage{algorithm}
\usepackage{algorithmic}
\usepackage{multirow}
\usepackage{graphicx, color}
\usepackage{booktabs}
\usepackage{epsfig}
\usepackage{epstopdf}
\usepackage{amsmath}
\usepackage{caption}
\usepackage{subfig}
\usepackage{amsfonts}
\usepackage{wasysym}
\usepackage{paralist}
\usepackage{url}
\usepackage{float}
\usepackage{booktabs}
\usepackage{balance}
\usepackage{marvosym}
\usepackage{ulem}
\newfloat{algorithm}{t}{lop}
\DeclareGraphicsExtensions{.pdf,.eps,.png,.jpg,.mps, .jpeg}

\newcommand{\ignore}[1]{}

\newcommand{\mar}[1]{\textcolor{magenta}{#1}}
\newcommand{\m}[1]{\textcolor{blue}{(#1)}}
\newcommand{\todo}[1]{\textcolor{red}{#1}}

\begin{document}

\title{Fitness Landscape Analysis of Dimensionally-Aware Genetic Programming Featuring Feynman Equations}

\titlerunning{FLA of DAGP Featuring Feynman Equations}

\author{Marko Durasevic \inst{1} 
\and
Domagoj Jakobovic\inst{1} \orcidID{0000-0002-9201-2994}
\and
Marcella Scoczynski Ribeiro Martins\inst{2}\orcidID{0000-0002-5716-4968} 
\and
Stjepan Picek\inst{3} 
\and
Markus Wagner\inst{4}\orcidID{0000-0002-3124-0061} 
}
\authorrunning{Durasevic et al.}
\institute{
University of Zagreb, Faculty of electrical engineering and computing, Croatia\\
\email{\{marko.durasevic, domagoj.jakobovic\}@fer.hr} 
\and
Federal University of Technology Paran\'a (UTFPR), Brazil\\
\email{marcella@utfpr.edu.br} 
\and
Delft University of Technology, The Netherlands\\
\email{s.picek@tudelft.nl}
\and
Optimisation and Logistics group, The University of Adelaide, Australia\\
\email{markus.wagner@adelaide.edu.au} 
}

\maketitle

\begin{abstract}
Genetic programming is an often-used technique for symbolic regression: finding symbolic expressions that match data from an unknown function.
To make the symbolic regression more efficient, one can also use dimensionally-aware genetic programming that constrains the physical units of the equation.
Nevertheless, there is no formal analysis of how much dimensionality awareness helps in the regression process.
In this paper, we conduct a fitness landscape analysis of dimensionally-aware genetic programming search spaces on a subset of equations from Richard Feynman's well-known lectures. 
We define an initialisation procedure and an accompanying set of neighbourhood operators for conducting the local search within the physical unit constraints.
Our experiments show that the added information about the variable dimensionality can efficiently guide the search algorithm.
Still, further analysis of the differences between the dimensionally-aware and standard genetic programming landscapes is needed to help in the design of efficient evolutionary operators to be used in a dimensionally-aware regression.

\keywords{Genetic programming \and dimensionally-aware GP \and fitness landscape \and local optima network}
\end{abstract}

\section{Introduction}
\label{sec:introduction}


Symbolic regression is 
a unique and very general type of multivariate regression analysis in which the task is to find the mathematical expression that links a number of variables in a domain with an unknown target function that would fit a dataset $S=\{(\mathbf{x^{(i)}}, y^{(i)})\}$, i.e., a set of pairs of an unknown multivariate target function $f : \mathbb{R}^n \to \mathbb{R}$.
With more than a quarter of a century of research in the field, the results obtained attracted the interests of many researchers to work in this area. A large number of applications of symbolic regression is both impressive, and it is also constantly expanding. For instance, symbolic regression has helped to extract physical laws using experimental data of chaotic dynamical systems without any knowledge of Newtonian mechanics~\cite{eureqa-article}. Others have used it to design more efficient antennas~\cite{koza} and to analyse satellite data~\cite{astronomy}. Symbolic regression via Genetic Programming (GP) implementations has been used to model mechanisms of drug response in cancer cell lines using genomics and experimental data~\cite{Fitzsimmons_Moscato_2018},
to discover hidden relationships in astronomical datasets~\cite{Graham_2013},
to predict wind farm output from weather data~\cite{wind},
to generate computer game scenes~\cite{DBLP:journals/ijcgt/FradeVC09},
for Boolean classification~\cite{DBLP:conf/eurogp/MuruzabalCF00}, and for many other scenarios.


In some sense, Evolutionary Computation (EC) methods for symbolic regression (most commonly employing GP-based implementations) somewhat ``compete'' with other strategies like support vector regression and artificial neural networks.
However, many researchers prefer to use symbolic regression since they tend to produce models with a significantly smaller number of variables, leading to solutions in a form amenable to downstream studies (e.g., uncertainty propagation and sensitivity analysis) and more ``explainable'' outcomes. 

Although symbolic regression methods -- and in particular GP-based methods -- are popular among some circles, the research has not sufficiently explored how to use problem-domain information, and even commercial products like Eureqa~\cite{eureqa-article} do not make use of it. With this paper, we propose to revisit the idea of Dimensionally-Aware Genetic Programming~\cite{10.5555/2934046.2934065} and to analyse the impact of design decisions using modern fitness landscape analysis tools.
To this end, we take a recent benchmark suite of symbolic regression problems~\cite{udrescu2019ai}, which also includes information about the dimensionality of input variables and the resulting model outputs.
Taking this information into account, we devise and employ a local search algorithm which at all times satisfies the imposed dimensionality constraints.
Using the local search, a complete network of local optima is built, considering given neighbourhood operators.
After the local optima network (LON) is obtained, information from the search is used to infer characteristics of the underlying fitness landscape.
At the same time, a comparison is made with the regular GP that does not restrict the dimensionality of the variables, to estimate the problem difficulty and the potential effectiveness of this approach.






\section{Background}
\label{sec:background}



\subsection{Feynman's Equations}

We will apply our methods to ``rediscover'' the fundamental physical laws. 
We consider equations from Feynman Lectures on Physics~\cite{Feynman:1494701}, covering topics like classical mechanics, electromagnetism, and quantum mechanics. 
Here, we follow the equation selection from Udrescu and Tegmark~\cite{udrescu2019ai}.
The authors listed 100 equations that do not contain derivatives or integrals and have between one and nine independent variables. 
The same authors also provide the Feynman Symbolic Regression Database~\cite{fdsr}, where for each equation, there is a data table whose rows are of the form ${x_1, x_2, \ldots y}$, where $y = f(x_1, x_2, \ldots)$. 
Table~\ref{tbl:equations} contains the 27 equations that we consider in the present paper. This subset was selected to involve equations with a varying number of variables, different types of operators, varying degrees of complexity, and a different number of physical units.
For the sake of readability, we will refer to these as the Feynman equations from now on.

\begin{table}[!h]
\centering
\vspace{-.5cm}
\caption{Feynman equations considered in this article; the units column shows the number of different physical units of the corresponding variables.}
\small
\label{tbl:equations}
\begin{tabular}{@{}lcccc@{}}
\toprule
ID & Feynman eq. & \multicolumn{1}{c}{Equation}                & \multicolumn{1}{c}{Variables} & \multicolumn{1}{c}{Units} \\ \midrule
1  & I.8.14      & d=$\sqrt[]{(x_2-x_1)^2+(y_2-y_1)^2}$        & 4                                       &   1                        \\
2  & I.12.1      & $F=\mu N_n$                                 & 2                                       &   3                       \\
3  & I.12.2      & $F=\frac{q_1 q_2}{4 \pi \epsilon r^2}$      & 4                                       &   4                       \\
4  & I.12.5      & $F=q_2 E_f$                                 & 2                                       &   4                        \\
5  & I.13.4      & $K=\frac{1}{2}m(v^2+u^2+w^2)$               & 4                                       &   3                       \\
6  & I.14.3      & $U=mgz$                                     & 3                                       &   3                        \\
7  & I.14.4      & $U=\frac{k_{spring}x^2}{2}$                 & 2                                       &   3                        \\
8  & I.18.4      & $r=\frac{m_1 r_1 + m_2 r_2}{m_1 + m_2}$     & 4                                       &   2                        \\
9  & I.24.6      & $E=\frac{1}{4}m(\omega^2+\omega^2_0)x^2$    & 4                                       &   3                        \\
10 & I.25.13     & $V_e=\frac{q}{C}$                           & 2                                       &   4                        \\
11 & I.27.6      & $f_f=\frac{1}{\frac{1}{d_1}+\frac{n}{d_2}}$ & 3                                       &   1                        \\
12 & I.29.4      & $k=\frac{\omega}{c}$                        & 2                                       &   2                        \\
13 & I.32.5      & $P=\frac{q^2a^2}{6 \pi \epsilon c^3}$       & 4                                       &   4                        \\
14 & I.34.8      & $\omega = \frac{qvB}{p}$                    & 4                                       &   4                        \\
15 & I.39.1      & $E_n = \frac{3}{2}pV$                      & 2                                       &     3                      \\
16 & I.39.22     & $P_F=\frac{nk_b T}{V}$                      & 4                                       &   4                        \\
17 & I.43.16     & $v=\frac{\mu q V_e}{d}$               & 4                                       &   4                        \\
18 & I.43.31     & $D=\mu_e k_b T$                             & 3                                       &   4                        \\
19 & II.2.42     & $P=\frac{\kappa (T_2 - T_1)A}{d}$          & 5                                       &    4                       \\
20 & II.8.31     & $E_{den}=\frac{\epsilon E^2_f}{2}$          & 2                                       &   4                        \\
21 & II.11.3     & $x=\frac{qE_f}{m(\omega^2_0-\omega^2)}$     & 5                                       &   4                        \\
22 & II.15.4     & $E=-\mu_M B \cos( \theta )$                 & 3                                       &   4                        \\
23 & II.34.2     & $\mu_M=\frac{qvr}{2}$                       & 3                                       &   4                        \\
24 & II.34.29b   & $E=\frac{g_{\_}\mu_M BJ_z}{h}$               & 5                                       &     4                      \\
25 & II.38.3     & $F=\frac{YAx}{d}$                           & 4                                       &   3                        \\
26 & III.13.18   & $v=\frac{2Ed^2k}{h}$                        & 4                                       &     3                      \\
27 & III.15.14   & $m=\frac{h^2}{2Ed^2}$                       & 3                                       &     3                      \\ \bottomrule
\end{tabular}
\vspace{-.5cm}
\end{table}


\subsection{Fitness Landscape Analysis}
\label{sec:fla}

Fitness landscapes illustrate the correlation between the search and fitness space~\cite{richter2014recent}, and are commonly used to describe or predict the performance of a heuristic search. 
Fitness landscape analysis can help predict the performance of heuristics by using search cost models. 
Local Optima Network (LON) is a fitness landscape model proposed in~\cite{ochoa2008study} for combinatorial landscapes, considering that the number and distribution of local optima in a search space represents an important impact on the performance of heuristic search algorithms~\cite{daolio2012local}. 
In this network model, the nodes are the local optima of a given optimisation problem, and the edges represent transitions among them using a neighbourhood operator \cite{verel2018sampling}. Therefore, the fitness landscape is represented as a graph of connected local optima.

In general, a local search heuristic $\mathbf{LS}$ maps the solution space $S$ to the set of locally optimal solutions $S^*$. A solution $i$ in the solution space $S$ is a local optimum given a neighbourhood operator $\mathcal{N}$ if $F(i)\geq F(s), \forall s \in \mathcal{N}(i)$. 
Each local optima $i$ has an associated basin of attraction corresponding to the set composed of all the solutions that, after applying the local search heuristic starting from each of them, the procedure returns $i$.
Therefore, the basin of attraction associated to a local optima $i$ is the set $B_{i}=\{s\in S | \mathbf{LS}(s)=i\}$ whose size is the cardinality of $B_{i}$.
In this paper, a connection (undirected) edge between two basins is created if at least one solution in one basin has a neighbour solution in the other basin, given a neighbourhood operator. This approach was also used in other works (e.g., \cite{ochoa2008study, yafrani2018fitness}).



\section{Technical Details}
\label{sec:details}

\subsection{Dimensionally-aware Genetic Programming}

The Dimensionally-Aware GP, first introduced by Keijzer and Babovic~\cite{10.5555/2934046.2934065}, can only be applied if there is information about the physical units of the model variables.
In~\cite{fdsr}, the authors provide the unit table that specifies the physical units of the input and output variables for all Feynman equations.
There are five different physical units appearing in all the equations: length $[m]$, time $[s]$, mass $[kg]$, temperature $[K]$, and potential $[V]$.
For every equation and each variable, the exact \textit{unit signature} is given. 
For instance, a variable denoting the distance is expressed in meters, and the corresponding signature would be $[1, 0, 0, 0, 0]$; a variable denoting acceleration is expressed in meters per second squared, and its signature can be presented with $[1, -2, 0, 0, 0]$.
Using the same notation, the result of each equation will have a corresponding \textit{target signature}. 
Following the dimensionally-aware paradigm, the local search algorithm we employ will always conform to the given target signature.
In other words, at all times, we only consider those candidate expressions that result in the desired signature.
Furthermore, when including the arithmetic operators in the expression, we follow the simple rules illustrated in Table~\ref{tab:effect_operations}: multiplication and division operators simply add or subtract the exponent values in the signature, while addition and subtraction can only be applied to expressions with the commensurate signature, and the resulting signature remains unchanged.

\begin{table}[!h]
\vspace{-.5cm}
\centering\small
        \caption{Effect of operations.}
        \label{tab:effect_operations}
        \begin{tabular}{p{2cm}p{8cm}}
            \toprule
            Function & Operations dimensionality \\ \midrule
            Addition & $[v,w,x,y,z],[v,w,x,y,z]\rightarrow[v,w,x,y,z]$ \\\midrule
            Subtraction & $[v,w,x,y,z],[v,w,x,y,z]\rightarrow[v,w,x,y,z]$ \\\midrule
            Multiplication & $[v,w,x,y,z],[v,w,x,y,z]\rightarrow[v+v,w+w,x+x,y+y,z+z]$ \\\midrule
            Division & $[v,w,x,y,z],[v,w,x,y,z]\rightarrow[v-v,w-w,x-x,y-y,z-z]$
            \\\bottomrule
        \end{tabular}
\vspace{-.5cm}
\end{table}

\subsection{Initialisation Procedure}

The goal of the initialisation procedure is to generate expressions whose result conforms to the target unit signature.
This is achieved by using all of the available variables and only multiplication and division operators.
In such an expression (e.g. $x^1y^{-2}z^0$), each variable can be represented only by its exponent, which is an integer value.
In initialisation, we consider exponents in the range $[-3,\ldots, 3]$; if $r$ is the cardinality of the range and if an equation has $p$ variables, this makes $r^p$ combinations to test. In the end, all combinations that yield the correct signature define the set of all possible initial solutions.
For instance, if the available variables represent time $t$ and distance $d$, and the target signature requires speed, the correct initial expressions would be $(t^{-1}d^1)$, $(t^{-2}d^2)$, etc.
Note, in the case where the chosen exponent range is not expressive enough to generate a single valid expression, the maximum exponent values can be increased and the initialisation simply restarted.

\subsection{Neighbourhood Operators}

\ignore{\todo{for the operators: mention that these are generic, but they seem to be new -- or something like this.}}

For our variation operators, we consider custom operators designed to be dimensionally-aware, i.e., their application does not change the signature of the overall expression encoded as a tree. 

\begin{compactitem}
\item \textbf{Replacement operator}. Select a subtree $t$ with a signature $s_t=[v,w,x,y,z]$ from the tree $T$ and replace it with a subtree $\hat{t}$ that has a commensurate signature, i.e., $s_t = s_{\hat{t}}$. 
\item \textbf{Multiplication with integer}. Select a subtree $t$ with a signature $s_t=[v,w,x,y,z]$ from the tree $T$ and replace it with a tree $\hat{t}$ where one child is $t$ and the other one is integer (dimensionless) in the range $[-3,\ldots, 3]$. The signatures of $t$ and $\hat{t}$ are the same.
\item \textbf{Divison with integer}. Same as the previous one, except the two subtrees are connected with the division operator.
\item \textbf{Addition with a commensurate value}. Select a subtree $t$ with a signature $s_t=[v,w,x,y,z]$ from the tree $T$ and replace it with a tree $\hat{t}$ where one child is $t$ and the other one is $q$ that has the same signature as $t$, i.e., $s_t = s_q$.
\item \textbf{Subtraction with a commensurate value}. Same as the previous one, except the two subtrees are connected with the subtraction operator.
\end{compactitem}

In all of the above operators, the new subtree is generated by following the same approach as in the initialisation procedure, enumerating all subtrees with the appropriate signature where the variable exponents are in the range $[-3,\ldots, 3]$.
This set of operators can produce expressions with only the four basic arithmetic operations; while executing the operations, the signatures of each subtree are updated according to the rules in Table~\ref{tab:effect_operations}.
In the local search procedure, we use all the neighbourhood operators to generate all possible neighbours, and only the one with the best fitness measure is retained.
Additionally, in the implementation, the maximum tree size is limited to 42 nodes, since with the repeated application of the same operator the expressions can bloat, i.e. achieve slightly smaller error values while the number of nodes becomes arbitrarily large.\footnote{We have experimented with a range of more open-ended bloat-control mechanism, e.g., lexicographic optimisation for fitness and size. However, we observed that even in our rather discrete setting, optimising I.8.14 or I.27.6 would result in trees of a size of over 256 nodes.}

Since the Feynman equations also contain constants in multiplication or addition operations, we additionally employ the \textit{linear scaling} technique~\cite{keijzer2003linear}.
With linear scaling, the original expression encoded as a tree $T$ is evaluated as $(a + b \cdot T)$; the coefficients $a$ and $b$ are determined by a simple linear regression where the sum of squared errors between the desired output and $(a + b \cdot T)$ is minimised.

\subsection{Local Search Procedure}

The local search used in our study is described in Algorithm~\ref{algo:ls}, where $\mathcal{N}(.)$ represents the neighbourhood of the given solution.
The algorithm is deterministic; if there are multiple solutions with the same fitness value within the neighbourhood, the algorithm will retain the first one that it encounters.
The local search is started using \textit{all} initial solutions obtained with initialisation to generate a LON for every considered equation.

\vspace{-0.5cm}
\begin{algorithm}[!h]
\small
\caption{A greedy local search heuristic\label{algo:ls}}
\begin{algorithmic}[1]
  \STATE $s \gets $ initial solution
  \WHILE{there is an improvement}
    \STATE $s^* = s$
    \FOR{each $s^{**}  \text{ in } \mathcal{N}(s)$}
        \IF{$F(s^{**}) > F(s^*)$}
          \STATE $s^* \gets s^{**}$
        \ENDIF
    \ENDFOR
    \STATE $s = s^*$
  \ENDWHILE
\end{algorithmic}
\end{algorithm}
\vspace{-0.5cm}

As the local search fitness measure, we use the mean squared error (\textit{MSE}) of the expression; a strict improvement is required for a new solution to be accepted.
The described local search with operators conforming to the dimensional constraints will be denoted as ``DAGP'' in the remainder of the text.

\ignore{
\m{Regarding strict improvement:} we can give this a try together with "MSE only". Why? This lack of bloat-control might be very counter-intuitive, but it might not be a problem after all as the MSE is still 
allows us to walk over plateaus, and this way allow us to explore certain modifications that would otherwise be very difficult to perform in a single mutation. Example: given an integer i first, do a sin() operation, then 
}

\subsection{Genetic Programming Regression}

Apart from the DAGP, we also applied a regular form of GP symbolic regression to the chosen set of equations.
The purpose of these GP experiments is to estimate the problem difficulty regarding the number of variables and complex dimensionality relations among the variables.
The GP regression is not concerned with physical units but is guided exclusively with the minimisation of \textit{MSE} given the training data.
In our experiments, the GP -- which is based on the GP package ECF~\cite{ECF} -- uses the same parameters for all considered equations, which are listed in Table~\ref{tbl:GP_parameters}.
The selection scheme is simple: in each iteration $k = 3$ individuals are selected at random, and the worst one is eliminated.
The remaining two are recombined to produce one offspring, which is then mutated with given individual mutation probability and returned to the population; both the crossover and the mutation type are chosen randomly in each invocation.

\begin{table}[!h]
\centering
\small
\vspace{-.9cm}
\caption{Genetic programming parameters.}
\label{tbl:GP_parameters}
\begin{tabular}{p{3.5cm}p{8cm}}
\toprule
\textbf{Parameter}       & \textbf{Value}                                            \\ \midrule
Population size          & 500                                                       \\
Function set             & +, -, *, /, sin, cos                                      \\
Individual mutation rate & 0.5                                                       \\
Tree max depth           & 6                                                         \\
Crossover operators      & subtree, one point, size fair, uniform, context preserved \\
Mutation operators       & subtree, hoist, node replace, permutation, shrink         \\
Termination criteria     & $10^5$ evaluations                                          \\
Number of runs           & 50                                                        \\ \bottomrule
\end{tabular}
\vspace{-1cm}
\end{table}

\ignore{
\section{To Decide}
\todo{PLEASE DELETE WHEN DEALT WITH:}
\begin{enumerate}
\item What ordering of local search operators and how many of the six to use? \m{discussed via email:(1)  each one in isolation, (2) all together in a random fixed permutation, (3) do approach (2) lots of times (requires averaging later)}
\item Enforce max tree size or not? \m{see above: strict improvement requirement, lexicographic improvement, ... }
\item what to do with negative numbers? or have "negation" instead as a function? \m{work with the integers above for now... negation as a function would be an alternative $-->$ essentially we need to do these combinations in the future}
\item what to report? number of successful searches (hits), average numbers of evaluations, average number of initial solutions, comparison with vanilla GP, average expression size, average LON characteristics, results depending on number of variables, influence of linear scaling
\end{enumerate}
}

\section{Results}
\label{sec:results}

In our experiments, we are considering the selected 27 Feynman's equations and apply the dimensionally-aware local search (DAGP) and a standard symbolic regression GP.
The number of data points for each equation was equal to 100, which were uniformly sampled from the available datasets~\cite{udrescu2019ai}.
The primary goal of DAGP is the exploration of the dimensionally-aware fitness landscape by building a corresponding LON for each equation.
The second goal is an estimate of the effort needed to successfully navigate such a landscape, in comparison with the standard symbolic regression.
In addition to the described DAGP configuration, we experimented with the following modifications: (a) reducing the integer constant range to $[-2,\ldots, 2]$ and $[-1,\ldots, 1]$; and (b) different operator ordering in local search (five permutations).
Furthermore, both the GP and all DAGP configurations were tested with and without the linear scaling.


\subsection{Algorithm Efficiency}

When considering the efficiency of the search, we define an acceptance criterion with the $\textit{MSE} < 10^{-9}$, i.e., a solution is considered ``correct'' (a hit) if its \textit{MSE} falls below this limit. 

Table~\ref{tbl:results} shows the number of evaluations needed to find a correct solution, while a dash denotes no such solution was found.
In the case of DAGP, these values are non-volatile since the local search procedure is deterministic.
In the case of GP, the number of evaluations needed is just an estimate; GP is executed 50 times, which either terminate after 100\,000 evaluations or when a correct solution is found. 
In case a solution is found in at least one run, the estimate is calculated as the total number of evaluations across all runs, divided by the number of successful runs (e.g., if each run was successful, this is equivalent to the average number of evaluations over all runs).

From the table, we can divide the equations into several groups; the first group are trivial problems, in which the dimensionally-aware approach needed very few evaluations to construct the correct solution.
In most cases, this is because the unit constraints result with only a single initial solution with the correct target signature.
The second group are the equations which are not trivial, but the DAGP can construct a correct solution using the local search operators and linear scaling.
For all these, the number of evaluations needed is considerably smaller than the corresponding GP search.

Finally, the third group includes equations which were not reconstructed; in some cases, this is because they include operators we have not considered, such as square root (I.18.14) or trigonometric functions (II.15.4).
The rest of those equations (I.13.4, I.18.4) also presented a challenge to the GP, since it was successful in a small number of runs requiring a large number of evaluations.
For both GP and DAGP, linear scaling was beneficial and provided improvement of the model, regardless of the representation.
It is also interesting to note that both DAGP modifications (a) and (b) made no difference in the number of equations whose solution was found, so we omit those results.
As an illustration, we applied the DAGP local search and GP with scaling to 39 additional equations from the benchmark (the ones not including trigonometric functions); the DAGP was able to find a solution for 28 equations, whereas GP converged in 29 cases.


\begin{table}[!h]
\centering
\caption{Number of evaluations needed to obtain the optimum. A value in brackets denotes the number of successful GP runs, '-' denotes unsuccessful run.}
\small
\label{tbl:results}
\begin{tabular}{@{}lrrrr@{}}
    \toprule
              & \multicolumn{2}{c}{DAGP local search} & \multicolumn{2}{c}{GP}       \\
    Eq. label & no scaling &                  scaling &      no scaling &    scaling \\ \midrule
    I.8.14    &          - &                        - &               - &          - \\
    I.12.1    &        267 &                      214 &        680 (50) &   620 (50) \\
    I.12.2    &          - &                        5 &               - &  1\,6750 (46) \\
    I.12.5    &          1 &                        1 &        580 (50) &   580 (50) \\
    I.13.4    &          - &                        - &               - &  2\,464\,750 (2) \\
    I.14.3    &          1 &                        1 &     2\,060 (50) &  2\,000 (50) \\
    I.14.4    &          - &                        1 &    908\,400 (5) &  1\,740 (50) \\
    I.18.4    &          - &                        - &    675\,785 (7) &  1\,613\,300 (3) \\
    I.24.6    &          - &                   2\,086 &               - &  2\,425\,250 (2) \\
    I.25.13   &          1 &                        1 &        960 (50) &   780 (50) \\
    I.27.6    &    72\,575 &                   2\,817 &   223\,735 (17) &  740\,500 (6) \\
    I.29.4    &          1 &                        1 &        950 (50) &   840 (50) \\
    I.32.5    &          - &                        1 &               - &  33\,370 (43) \\
    I.34.8    &          1 &                        1 &    20\,076 (46) &  4\,620 (50) \\
    I.39.1    &          - &                        1 & 1\,574\,500 (3) &   560 (50) \\
    I.39.22   &        517 &                      408 &    15\,904 (47) &  4\,800 (50) \\
    I.43.16   &          1 &                        1 &    21\,488 (45) &  6\,260 (50) \\
    I.43.31   &          1 &                        1 &     2\,080 (50) &  2\,110 (50) \\
    II.2.42   &    19\,468 &                  29\,556 &    98\,450 (30) & 22\,500 (48) \\
    II.8.31   &          - &                        1 & 1\,155\,625 (4) &  1\,760 (50) \\
    II.11.3   &     1\,000 &                   2\,042 & 4\,921\,500 (1) &  940\,000 (5) \\
    II.15.4   &          - &                        - &    43\,397 (39) &  3\,750 (50) \\
    II.34.2   &          - &                        1 & 1\,161\,875 (4) &  1\,820 (50) \\
    II.34.29b &          - &                   4\,355 &               - &  8\,400 (50) \\
    II.38.3   &        120 &                      120 &    11\,030 (49) &  4\,100 (50) \\
    III.13.18 &          - &                       45 &               - &  6\,400 (50) \\
    III.15.14 &          - &                        1 &               - &  10\,950 (48) \\ \bottomrule
\end{tabular}
\vspace{-0.7cm}
\end{table}

\subsection{LON characteristics for DAGP}



\begin{table}[!h]
\small
\vspace{-.2cm}
\centering
\caption{Graph metrics for DAGP local search.}
\label{tab:statistics0405}
\small
\begin{tabular}{lllllllll|lllllllll}
\toprule
 & 
\multicolumn{8}{c}{no-scaling} & \multicolumn{8}{c}{linear-scaling}\\    
\toprule
\textbf{equation}& 
\textbf{$n_v$} & 
\textbf{$n_e$} & 
\textbf{$\overline{C}$} & 
\textbf{$\overline{C_r}$} & 
\textbf{$\overline{l}$} & 
\textbf{$\pi$} & 
\textbf{$S$} & 
\textbf{$n_\textit{hits}$} &

\textbf{$n_v$} & 
\textbf{$n_e$} & 
\textbf{$\overline{C}$} & 
\textbf{$\overline{C_r}$} & 
\textbf{$\overline{l}$} & 
\textbf{$\pi$} & 
\textbf{$S$} & 
\textbf{$n_\textit{hits}$} \\ \midrule
I.8.14    &$    220    $&$    1641    $&$    0.85    $&$    0.07    $&$    -1.00    $&$    0    $&$    17    $&$    0    $&$    223    $&$    1805    $&$    0.87    $&$    0.07    $&$    -1.00    $&$    0    $&$    6    $&$    0    $\\
I.12.1    &$    5    $&$    4    $&$    0.47    $&$    0.00    $&$    -1.00    $&$    0    $&$    2    $&$    5    $&$    3    $&$    2    $&$    0.00    $&$    0.00    $&$    1.33    $&$    1    $&$    1    $&$    3    $\\
I.12.2    &$    5    $&$    6    $&$    0.80    $&$    0.53    $&$    -1.00    $&$    0    $&$    2    $&$    0    $&$    5    $&$    6    $&$    0.80    $&$    0.53    $&$    -1.00    $&$    0    $&$    2    $&$    5    $\\
I.12.5    &$    1    $&$    0    $&$    0.00    $&$    0.00    $&$    0.00    $&$    1    $&$    1    $&$    1    $&$    1    $&$    0    $&$    0.00    $&$    0.00    $&$    0.00    $&$    1    $&$    1    $&$    1    $\\
I.13.4    &$    32    $&$    66    $&$    0.84    $&$    0.15    $&$    -1.00    $&$    0    $&$    6    $&$    0    $&$    33    $&$    67    $&$    0.80    $&$    0.18    $&$    -1.00    $&$    0    $&$    6    $&$    0    $\\
I.14.3    &$    1    $&$    0    $&$    0.00    $&$    0.00    $&$    0.00    $&$    1    $&$    1    $&$    1    $&$    1    $&$    0    $&$    0.00    $&$    0.00    $&$    0.00    $&$    1    $&$    1    $&$    1    $\\
I.14.4    &$    1    $&$    0    $&$    0.00    $&$    0.00    $&$    0.00    $&$    1    $&$    1    $&$    1    $&$    1    $&$    0    $&$    0.00    $&$    0.00    $&$    0.00    $&$    1    $&$    1    $&$    1    $\\
I.18.4    &$    41    $&$    73    $&$    0.77    $&$    0.03    $&$    -1.00    $&$    0    $&$    11    $&$    0    $&$    42    $&$    72    $&$    0.80    $&$    0.06    $&$    -1.00    $&$    0    $&$    11    $&$    0    $\\
I.24.6    &$    5    $&$    10    $&$    1.00    $&$    1.00    $&$    1.00    $&$    1    $&$    1    $&$    0    $&$    5    $&$    10    $&$    1.00    $&$    1.00    $&$    1.00    $&$    1    $&$    1    $&$    4    $\\
I.27.6    &$    39    $&$    100    $&$    0.61    $&$    0.09    $&$    -1.00    $&$    0    $&$    6    $&$    3    $&$    41    $&$    100    $&$    0.58    $&$    0.10    $&$    -1.00    $&$    0    $&$    8    $&$    25    $\\
I.29.4    &$    1    $&$    0    $&$    0.00    $&$    0.00    $&$    0.00    $&$    1    $&$    1    $&$    1    $&$    1    $&$    0    $&$    0.00    $&$    0.00    $&$    0.00    $&$    1    $&$    1    $&$    1    $\\
I.32.5    &$    1    $&$    0    $&$    0.00    $&$    0.00    $&$    0.00    $&$    1    $&$    1    $&$    0    $&$    1    $&$    0    $&$    0.00    $&$    0.00    $&$    0.00    $&$    1    $&$    1    $&$    1    $\\
I.34.8    &$    1    $&$    0    $&$    0.00    $&$    0.00    $&$    0.00    $&$    1    $&$    1    $&$    1    $&$    1    $&$    0    $&$    0.00    $&$    0.00    $&$    0.00    $&$    1    $&$    1    $&$    1    $\\
I.39.1    &$    1    $&$    0    $&$    0.00    $&$    0.00    $&$    0.00    $&$    1    $&$    1    $&$    0    $&$    1    $&$    0    $&$    0.00    $&$    0.00    $&$    0.00    $&$    1    $&$    1    $&$    1    $\\
I.25.13    &$    1    $&$    0    $&$    0.00    $&$    0.00    $&$    0.00    $&$    1    $&$    1    $&$    1    $&$    1    $&$    0    $&$    0.00    $&$    0.00    $&$    0.00    $&$    1    $&$    1    $&$    1    $\\
I.39.22    &$    6    $&$    6    $&$    1.00    $&$    0.44    $&$    -1.00    $&$    0    $&$    2    $&$    6    $&$    7    $&$    9    $&$    1.00    $&$    0.21    $&$    -1.00    $&$    0    $&$    2    $&$    7    $\\
I.43.16    &$    1    $&$    0    $&$    0.00    $&$    0.00    $&$    0.00    $&$    1    $&$    1    $&$    1    $&$    1    $&$    0    $&$    0.00    $&$    0.00    $&$    0.00    $&$    1    $&$    1    $&$    1    $\\
I.43.31    &$    1    $&$    0    $&$    0.00    $&$    0.00    $&$    0.00    $&$    1    $&$    1    $&$    1    $&$    1    $&$    0    $&$    0.00    $&$    0.00    $&$    0.00    $&$    1    $&$    1    $&$    1    $\\
II.2.42    &$    24    $&$    57    $&$    0.96    $&$    0.14    $&$    -1.00    $&$    0    $&$    4    $&$    3    $&$    23    $&$    49    $&$    0.92    $&$    0.14    $&$    -1.00    $&$    0    $&$    4    $&$    2    $\\
II.8.31    &$    1    $&$    0    $&$    0.00    $&$    0.00    $&$    0.00    $&$    1    $&$    1    $&$    1    $&$    1    $&$    0    $&$    0.00    $&$    0.00    $&$    0.00    $&$    1    $&$    1    $&$    1    $\\
II.11.3    &$    4    $&$    0    $&$    0.00    $&$    0.00    $&$    -1.00    $&$    0    $&$    4    $&$    4    $&$    5    $&$    1    $&$    0.00    $&$    0.00    $&$    -1.00    $&$    0    $&$    4    $&$    3    $\\
II.15.4    &$    7    $&$    6    $&$    0.86    $&$    0.21    $&$    -1.00    $&$    0    $&$    3    $&$    0    $&$    5    $&$    3    $&$    0.60    $&$    0.00    $&$    -1.00    $&$    0    $&$    3    $&$    0    $\\
II.34.2    &$    1    $&$    0    $&$    0.00    $&$    0.00    $&$    0.00    $&$    1    $&$    1    $&$    1    $&$    1    $&$    0    $&$    0.00    $&$    0.00    $&$    0.00    $&$    1    $&$    1    $&$    1    $\\
II.38.3    &$    13    $&$    10    $&$    0.64    $&$    0.00    $&$    -1.00    $&$    0    $&$    6    $&$    12    $&$    14    $&$    11    $&$    0.60    $&$    0.00    $&$    -1.00    $&$    0    $&$    6    $&$    14    $\\
II.34.29b    &$    46    $&$    250    $&$    0.76    $&$    0.24    $&$    -1.00    $&$    0    $&$    3    $&$    0    $&$    39    $&$    238    $&$    0.87    $&$    0.33    $&$    -1.00    $&$    0    $&$    5    $&$    36    $\\
III.13.18    &$    6    $&$    15    $&$    1.00    $&$    1.00    $&$    1.00    $&$    1    $&$    1    $&$    0    $&$    3    $&$    3    $&$    1.00    $&$    1.00    $&$    1.00    $&$    1    $&$    1    $&$    3    $\\
III.15.14    &$    1    $&$    0    $&$    0.00    $&$    0.00    $&$    0.00    $&$    1    $&$    1    $&$    0    $&$    1    $&$    0    $&$    0.00    $&$    0.00    $&$    0.00    $&$    1    $&$    1    $&$    1    $\\
\bottomrule                            
\end{tabular}
\vspace{-.5cm}
\end{table}


We expand the analysis extracting LONs from both DAGP landscapes, linear and no-scaling strategies.
The obtained networks can be analysed according to some general graph metrics useful to understand the landscape behaviour. Table~\ref{tab:statistics0405} reports the considered metrics: $n_v$ and $n_e$ represent the number of vertices (or nodes) and the number of edges of the generated LON, respectively. 
$C$ is the average clustering coefficient which measures cliquishness of a neighbourhood, and it characterises the degree to which nodes in a graph tend to cluster together; 
$C_r$ is the average clustering coefficient of corresponding random graphs (i.e., random graphs with the same number of vertices and mean degree).
$l$ is the average shortest path length between any two local optima. 
$\pi$ is the connectivity, which indicates if the LON is a connected graph with $S$ being the number of connected components (sub-graphs). 
Finally, $n_\textit{hits}$ is the number of nodes which represent a hit; as before, we consider a solution to be a hit if its mean square error is $\textit{MSE}<10^{-9}$.
Some landscapes (13 of the 27 reported in Table~\ref{tab:statistics0405}) consist of only a single node. Within the non-scaling experiments, the optimum appears in seven of these 13 cases; for linear scaling, the optimum is found in all 13 landscapes with unique nodes.

\begin{figure}[!h]
\vspace{.2cm}
\centering
\rotatebox{90}{\hspace{2mm}\footnotesize{\textit{no-scaling}}}\rotatebox{90}{\rule{20mm}{0.3pt}}%
\subfloat[I.24.6]{
  \centering
  \includegraphics[scale=0.2,clip,trim=40 40 40 40]{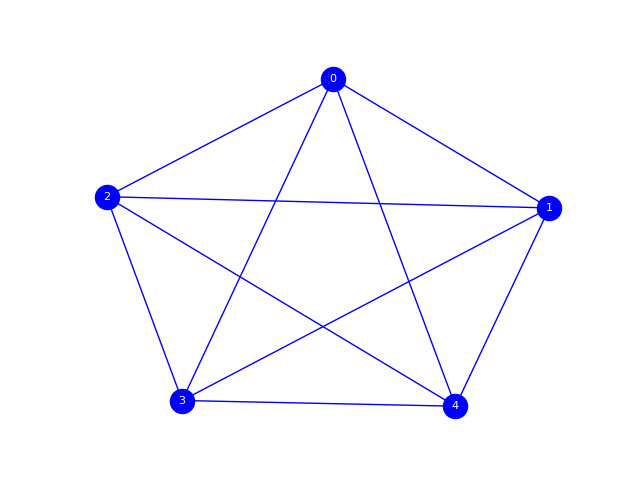}
  \label{fig:04I246}
}
\subfloat[III.13.18]{
  \centering
  \includegraphics[scale=0.2,clip,trim=40 40 40 40]{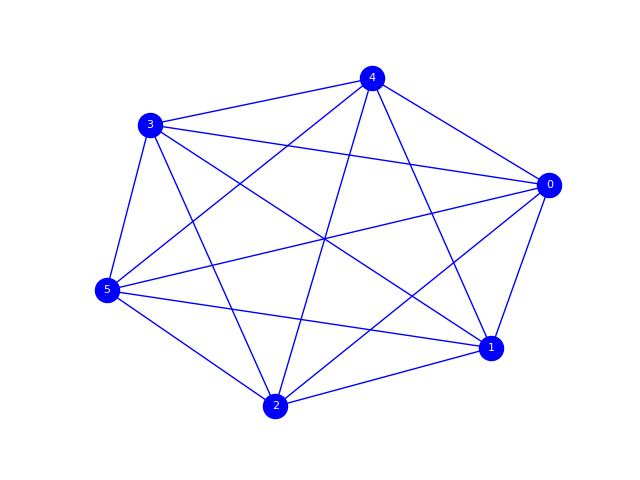}
  \label{fig:04III1318}
}
\rotatebox{90}{\hspace{1mm}\footnotesize{\textit{linear scaling}}}\rotatebox{90}{\rule{20mm}{0.3pt}}%
\subfloat[I.24.6]{
  \centering
  \includegraphics[scale=0.2,clip,trim=40 40 40 40]{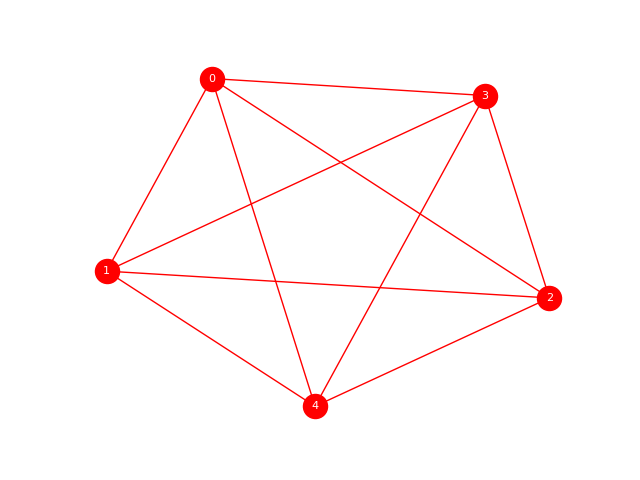}
  \label{fig:05I246}
}
\subfloat[III.13.18]{
  \centering
  \includegraphics[scale=0.2,clip,trim=40 40 40 40]{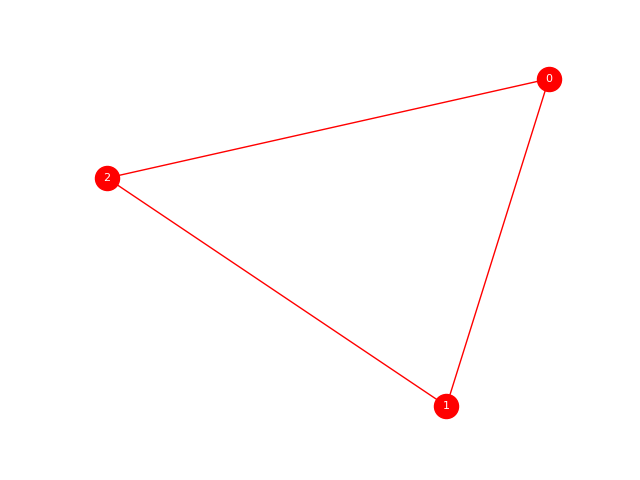}
  \label{fig:05III1318}
}
\vspace{-.2cm}
\caption{LON examples of fully-connected networks using no-scaling (lest-blue) and linear scaling (right-red) for I.24.6 and II.13.18.}\vspace{-.2cm}
\label{fig:lons_lbar}
\end{figure}

Analysing the average shortest path lengths ($\overline{l}$), some results show that the network is weakly and sometimes not connected ($\overline{l}=-1$). Few reported cases present $\overline{l}\geq1$, i.e., any pair of local optima can be connected by traversing at least other local optima, such as in I.24.6  and III.13.18 $\overline{l}=1$. Besides, in these examples, $\pi=1$ and $S=1$, meaning the network is connected in one entire component (see Figure~\ref{fig:lons_lbar} for examples).

\begin{figure}[!h]
\centering
\rotatebox{90}{\hspace{2mm}\small{\textit{no-scaling}}}\rotatebox{90}{\rule{20mm}{0.3pt}}%
\subfloat[I.13.4]{
  \centering
  \includegraphics[scale=0.2,clip,trim=40 40 40 40]{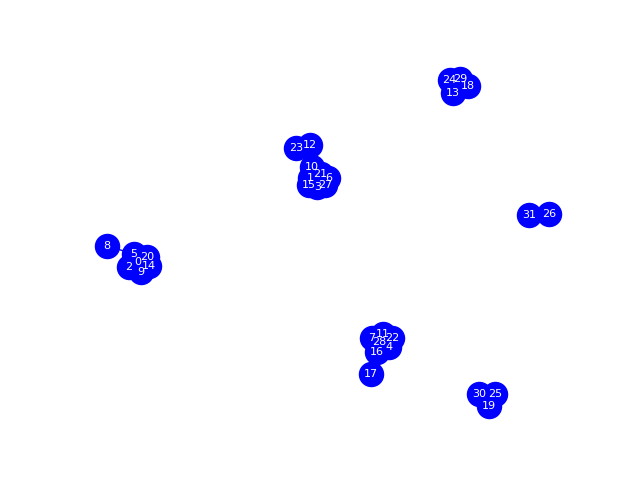}
  \label{fig:04I134}
}
\subfloat[I.18.4]{
  \centering
  \includegraphics[scale=0.2,clip,trim=40 40 40 40]{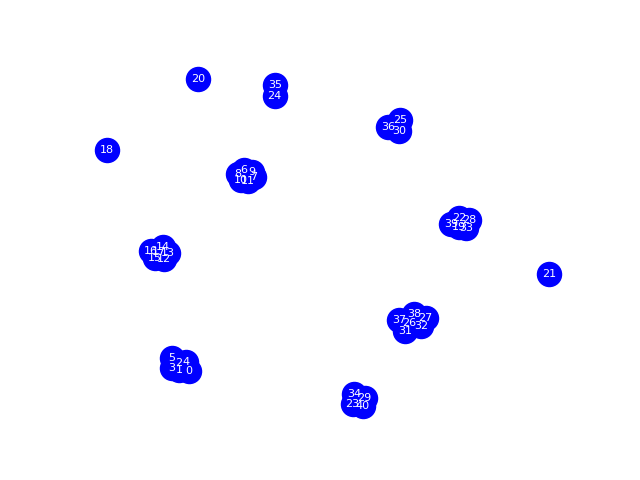}
  \label{fig:04I134}
}
\rotatebox{90}{\hspace{1mm}\small{\textit{linear-scaling}}}\rotatebox{90}{\rule{20mm}{0.3pt}}%
\subfloat[I.13.4]{
  \centering
  \includegraphics[scale=0.2,clip,trim=40 40 40 40]{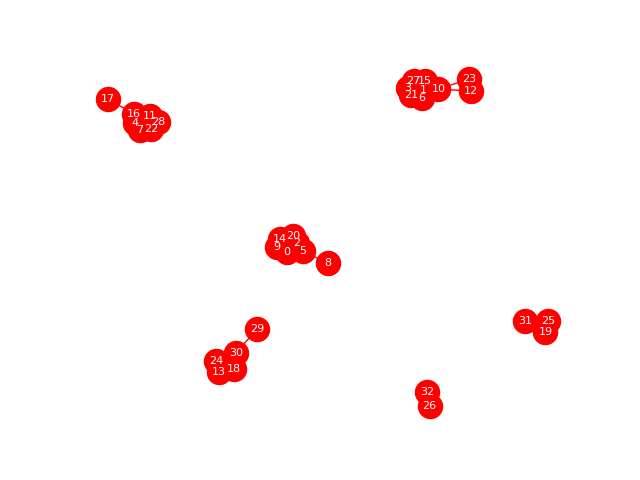}
  \label{fig:05I134}
}
\subfloat[I.18.4]{
  \centering
  \includegraphics[scale=0.2,clip,trim=40 40 40 40]{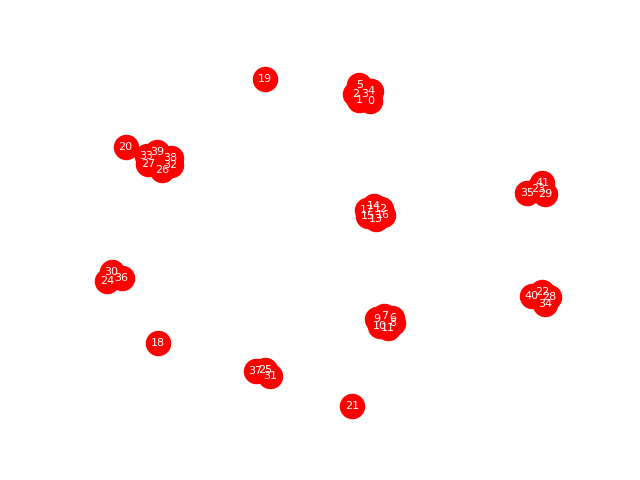}
  \label{fig:05I134}
}
\vspace{-.2cm}
\caption{LON examples of dense local clusters using no-scaling (left-blue) and linear scaling (right-red) for  I.13.4 and I.18.4.}
\label{fig:lons_cluster}
\end{figure}

We can also observe small-world properties by looking at the clustering coefficients ($\overline{C}$, $\overline{C_r}$) for some equations. 
Some LONs show a significantly high degree of local clustering compared with their corresponding random graphs, meaning that the local optima are connected in two ways: dense local clusters and sparse interconnections, which can be challenging to find and exploit (see examples in Figure~\ref{fig:lons_cluster} for I.13.4 and I.18.4).

\begin{figure}[!h]
\centering
\rotatebox{90}{\hspace{1mm}\small{\textit{no-scaling}}}\rotatebox{90}{\rule{17mm}{0.3pt}}%
\quad 
\subfloat[I.12.1]{
  \centering
  \includegraphics[scale=0.14]{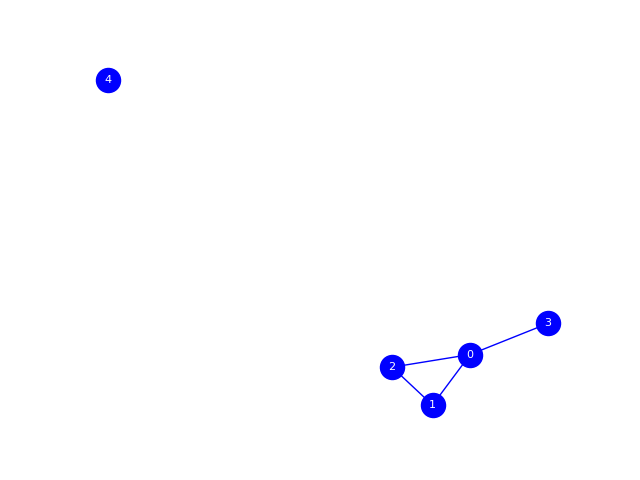}
  \label{fig:04I121}
}
\quad 
\subfloat[I.27.6]{
  \centering
  \includegraphics[scale=0.14]{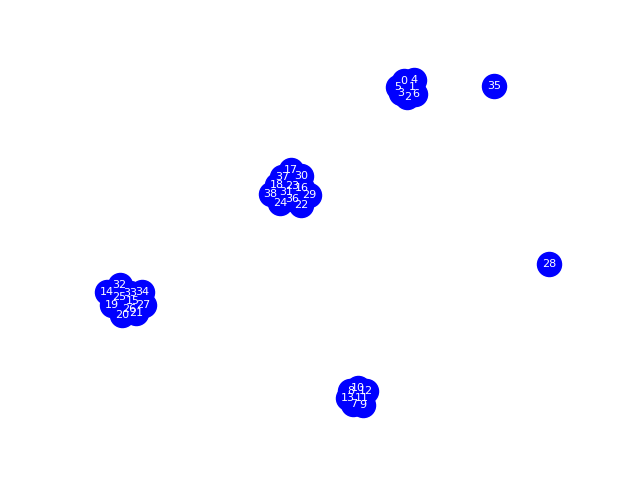}
  \label{fig:04I276}
}
\quad 
\subfloat[I.13.4]{
  \centering
  \includegraphics[scale=0.14]{./figures/04I134.png}
  \label{fig:04I134}
}
\quad 
\subfloat[II.34.29b]{
  \centering
  \includegraphics[scale=0.14]{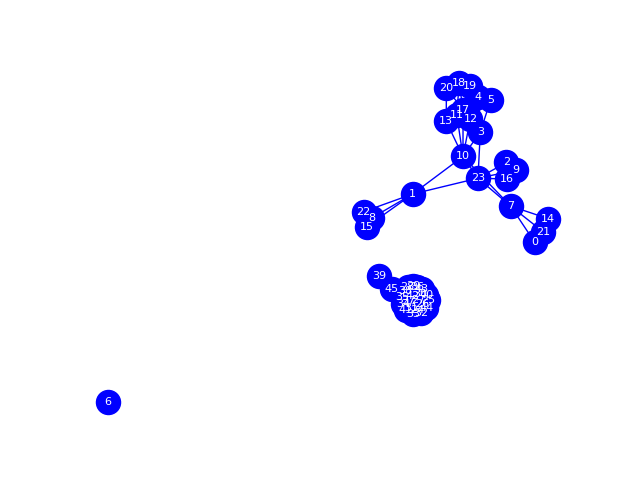}
  \label{fig:04II3429b}
}\\
\rotatebox{90}{\hspace{0mm}\small{\textit{linear-scaling}}}\rotatebox{90}{\rule{19mm}{0.3pt}}%
\quad
\subfloat[I.12.1]{
  \centering
  \includegraphics[scale=0.14]{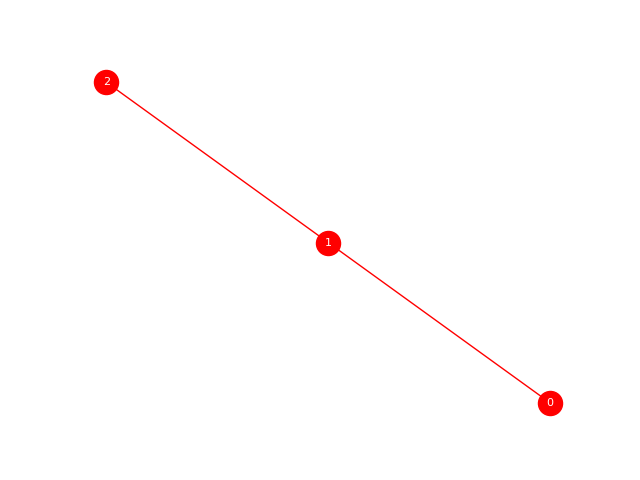}
 \label{fig:05I121}
}
\quad 
\subfloat[I.27.6]{
  \centering
  \includegraphics[scale=0.14]{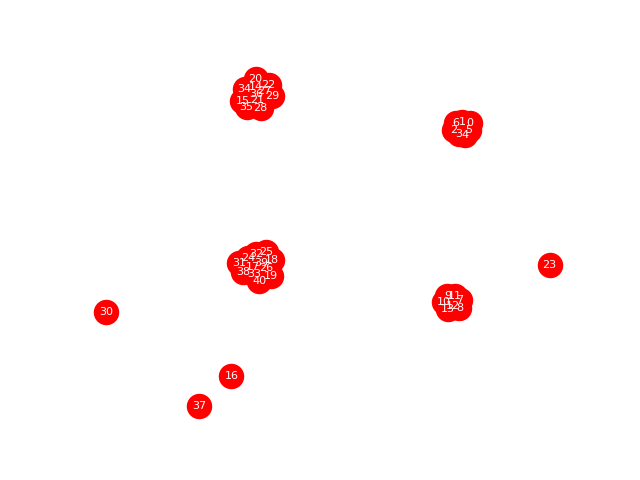}
  \label{fig:05I276}
}
\quad 
\subfloat[I.13.4]{
  \centering
  \includegraphics[scale=0.14]{./figures/05I134.png}
  \label{fig:05I134}
}
\quad 
\subfloat[II.34.29b]{
  \centering
  \includegraphics[scale=0.14]{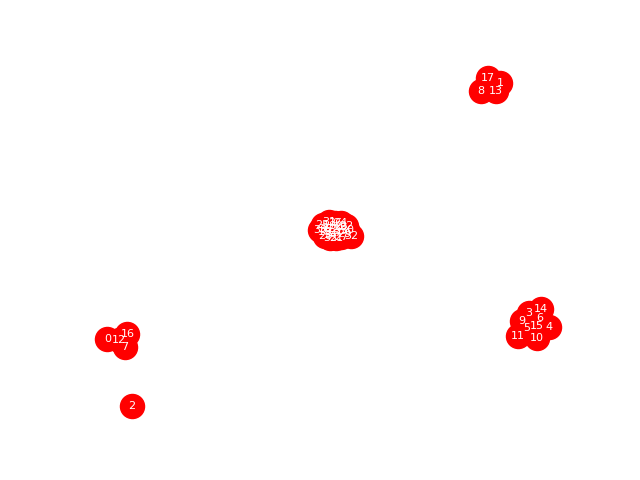}
  \label{fig:05II3429b}
}
\vspace{-.2cm}
\caption{LON using no-scaling (top-blue) and linear scaling (bottom-red) for particular equation examples with 2 (I.12.1), 3 (I.27.6), 4 (I.8.14) and 5 (II.34.29b) variables.}
\label{fig:lons_variables}
\vspace{-0.3cm}
\end{figure}

In Figure~\ref{fig:lons_variables}, we highlight particular LON examples with two (I.12.1), three (I.27.6), four (I.8.14), and five (II.34.29b) variables. Note that the $\overline{C}$ coefficient is higher for linear scaling in II.34.29b in comparison with no-scaling. Moreover, I.12.1 and I.27.6 present $n_\textit{hits}>0$; this also happens for II.34.29b but only considering linear scaling $n_\textit{hits}=36$.

\begin{figure}[!h]
\vspace{-0.3cm}
\centering
\subfloat[$n_v$]{
  \centering
  \includegraphics[scale=0.15]{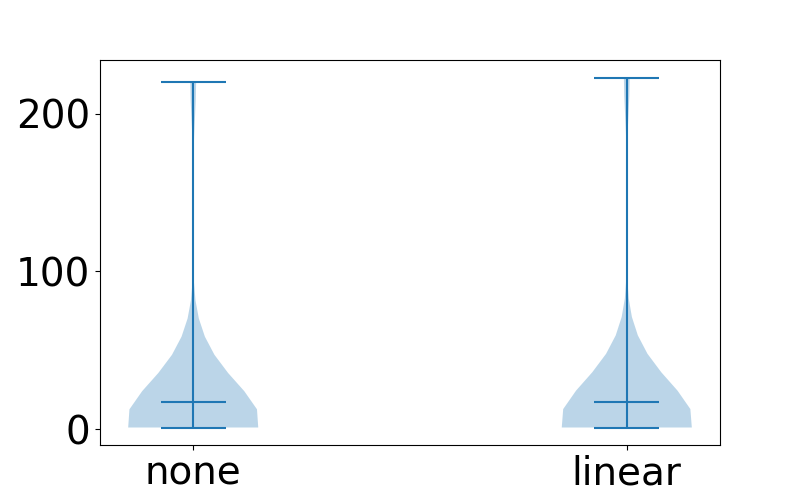}
  \label{fig:violin_nv}
}
\quad 
\subfloat[$n_e$]{
  \centering
  \includegraphics[scale=0.15]{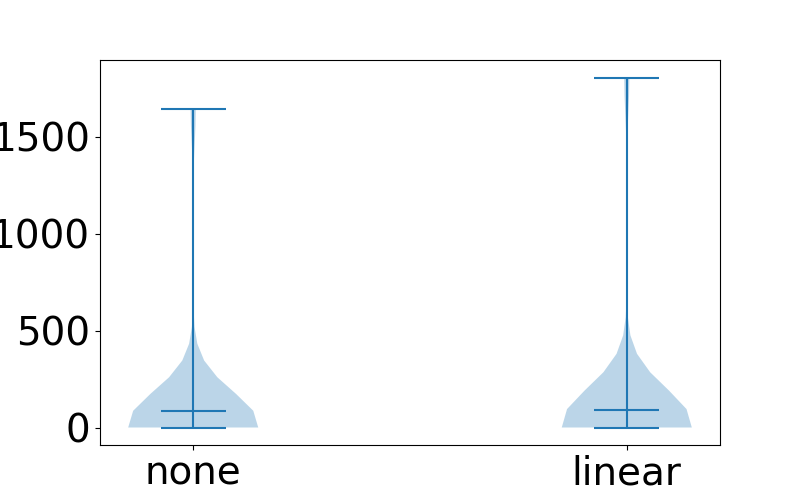}
  \label{fig:violin_ne}
}
\quad 
\subfloat[$\overline{C}$]{
  \centering
  \includegraphics[scale=0.15]{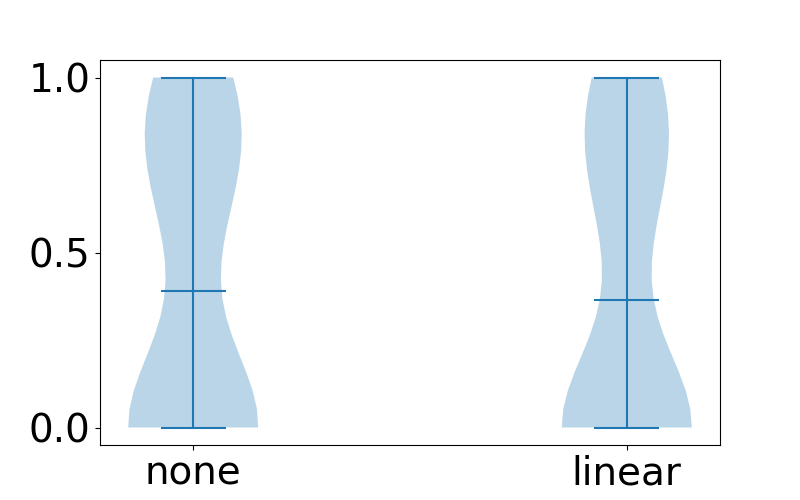}
  \label{fig:violin_C}
}
\quad 
\subfloat[$\overline{C_r}$]{
  \centering
  \includegraphics[scale=0.15]{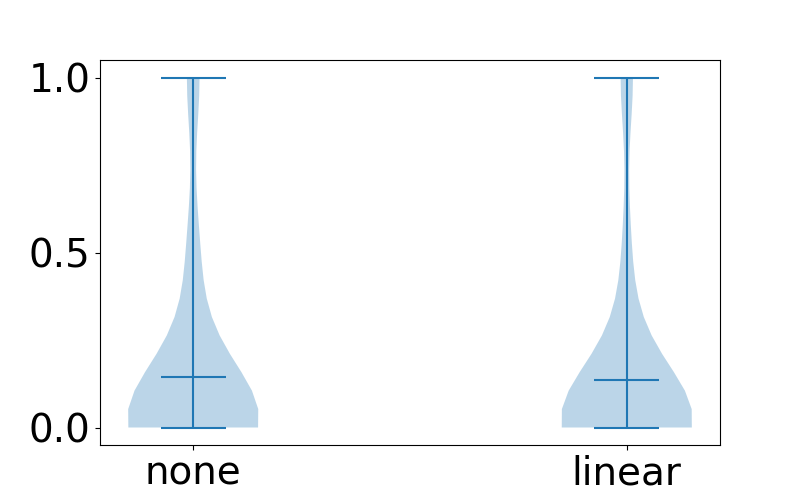}
  \label{fig:violin_Cr}
}
\subfloat[$\overline{l}$]{
  \centering
  \includegraphics[scale=0.15]{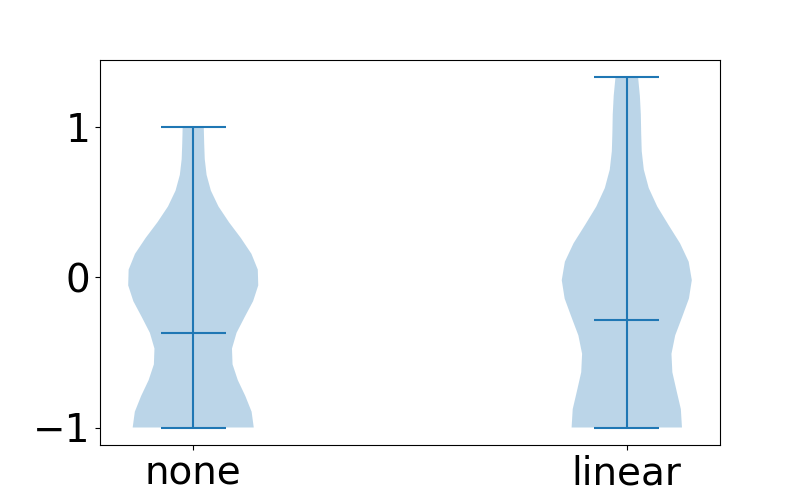}
  \label{fig:violin_l}
}
\quad 
\subfloat[$\pi$]{
  \centering
  \includegraphics[scale=0.15]{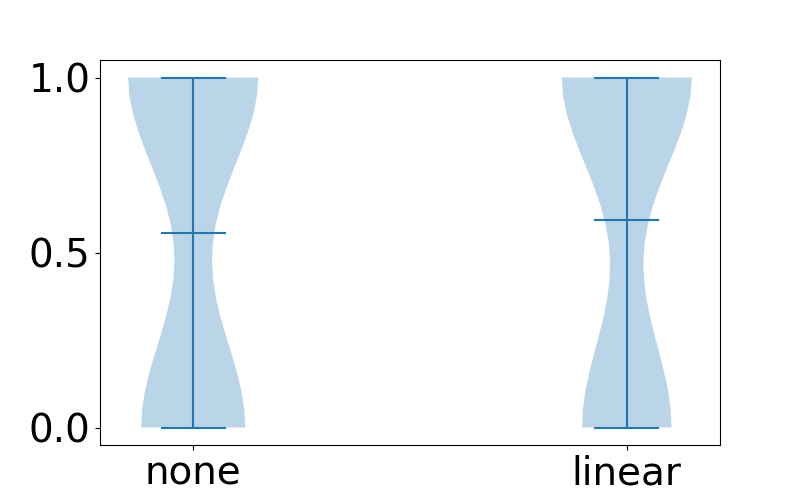}
  \label{fig:violin_pi}
}
\quad 
\subfloat[$S$]{
  \centering
  \includegraphics[scale=0.15]{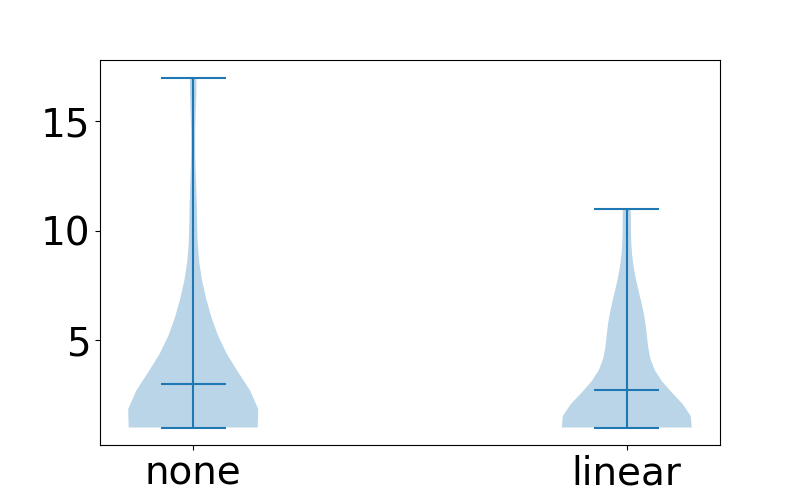}
  \label{fig:violin_S}
}
\quad 
\subfloat[$n_\textit{hits}$]{
  \centering
  \includegraphics[scale=0.15]{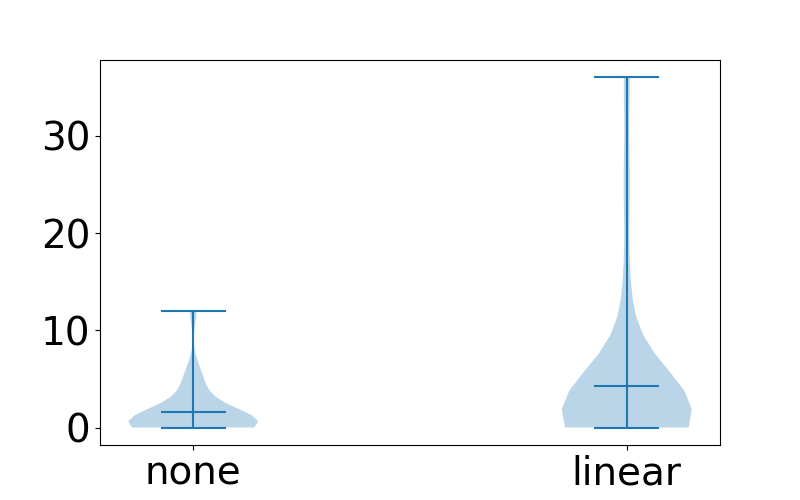}
  \label{fig:violin_nhits}
}
\vspace{-.2cm}
\caption{Violin plots for each graph metric over all equations.
The bar in the center represents the mean while the extremes denote upper and lower bounds.}
\label{fig:violin-run0405}
\vspace{-0.3cm}
\end{figure}

Figure~\ref{fig:violin-run0405} summarises each metric considering all addressed equations for both cases no-scaling and linear scaling. We note that with few exceptions ($\overline{l}$ and $n_\textit{hits}$), the metrics present similar distributions for both strategies. 
Since the two DAGP modifications (a) and (b) exhibit very similar behaviour, their graph metrics are not included.


\section{Conclusions and Future Work}
\label{sec:conclusion}

In many regression problems, only the raw data, obtained with the help of some measurements, is available to infer the governing model. 
It is not often the case that the information about the physical units of the result and the variables are documented; however, if this information is available, it can significantly improve regression to the extent that some problems become trivial to solve with the right approach.

Our experiments on a subset of equations of Richard Feynman's have shown that a very simple local search procedure, adhering to the dimensionally-aware constraints, can efficiently navigate the corresponding landscape and arrive at the correct solution. However, it must be noted that in real-world situations, a certain amount of noise in the data can be expected, which was not present in this study.

We also have extracted Local Optima Networks (LONs) providing a fitness landscape analysis for the dimensionally-aware genetic programming search space. The networks presented small-world properties for some equations meaning that the local optima can be connected as dense local clusters but also in sparse interconnections -- and sparse interconnections might make the search process harder even using strategies such as linear scaling. 

We plan to extend the dimensionally-aware local search to cover additional operators such as square root, exponential and trigonometric functions.
Besides local search, experiments can be performed by incorporating DA constraints into the standard GP, with appropriate mutation and crossover operators, where different fitness landscape models can be applied.
At the same time, a transition from the regular GP and DAGP could be achieved with the use of maximum deviation to the target signature, which could be gradually decreased over the course of the evolution.

\bibliographystyle{splncs04}
\bibliography{references}

\begin{thebibliography}{10}
\providecommand{\url}[1]{\texttt{#1}}
\providecommand{\urlprefix}{URL }
\providecommand{\doi}[1]{https://doi.org/#1}

\bibitem{ECF}
Evolutionary computation framework (2019), \url{http://ecf.zemris.fer.hr/}

\bibitem{daolio2012local}
Daolio, F., Verel, S., Ochoa, G., Tomassini, M.: Local optima networks and the
  performance of iterated local search. In: Genetic and Evolutionary
  Computation Conference (GECCO). pp. 369--376. ACM (2012)

\bibitem{Feynman:1494701}
Feynman, R.P., Leighton, R.B., Sands, M.: {The Feynman lectures on physics; New
  millennium ed.} Basic Books, New York, NY (2010),
  \url{https://cds.cern.ch/record/1494701}, originally published 1963-1965

\bibitem{Fitzsimmons_Moscato_2018}
Fitzsimmons, J., Moscato, P.: Symbolic regression modelling of drug responses.
  In: First IEEE Conference on Artificial Intelligence for Industries (2018)

\bibitem{DBLP:journals/ijcgt/FradeVC09}
Frade, M., de~Vega, F.F., Cotta, C.: Breeding terrains with genetic terrain
  programming: The evolution of terrain generators. Computer Games Technology
  \textbf{2009},  125714:1--125714:13 (2009)

\bibitem{astronomy}
Graham, M.J., Djorgovski, S.G., Mahabal, A., Donalek, C., Drake, A., Longo, G.:
  Data challenges of time domain astronomy. Distributed and Parallel Databases
  \textbf{30}(5),  371--384 (Oct 2012)

\bibitem{Graham_2013}
Graham, M., Djorgovski, S., Mahabal, A., Donalek, C., Drake, A.:
  Machine-assisted discovery of relationships in astronomy. Monthly Notices of
  the Royal Astronomical Society  \textbf{431}(3),  2371--2384 (2013)

\bibitem{keijzer2003linear}
Keijzer, M.: Improving symbolic regression with interval arithmetic and linear
  scaling. In: European Conference on Genetic Programming (EuroGP). pp. 70--82.
  Springer (2003)

\bibitem{10.5555/2934046.2934065}
Keijzer, M., Babovic, V.: Dimensionally aware genetic programming. In: 1st
  Annual Conference on Genetic and Evolutionary Computation (GECCO) Volume 2.
  p. 1069–1076. Morgan Kaufmann Publishers Inc. (1999)

\bibitem{koza}
Koza, J.R.: Genetic Programming: On the Programming of Computers by Means of
  Natural Selection. MIT Press (1992)

\bibitem{DBLP:conf/eurogp/MuruzabalCF00}
Muruz{\'a}bal, J., Cotta-Porras, C., Fern{\'a}ndez, A.: Some probabilistic
  modelling ideas for boolean classification in genetic programming. In: Poli,
  R., Banzhaf, W., Langdon, W.B., Miller, J., Nordin, P., Fogarty, T.C. (eds.)
  Genetic Programming. pp. 133--148. Springer (2000)

\bibitem{ochoa2008study}
Ochoa, G., Tomassini, M., V{\'e}rel, S., Darabos, C.: {A study of NK
  landscapes' basins and local optima networks}. In: Genetic and Evolutionary
  Computation Conference (GECCO). pp. 555--562. ACM (2008)

\bibitem{richter2014recent}
Richter, H., Engelbrecht, A.: Recent advances in the theory and application of
  fitness landscapes. Springer (2014)

\bibitem{eureqa-article}
Schmidt, M., Lipson, H.: Distilling free-form natural laws from experimental
  data. Science  \textbf{324}(5923),  81--85 (2009)

\bibitem{udrescu2019ai}
Udrescu, S.M., Tegmark, M.: Ai feynman: a physics-inspired method for symbolic
  regression (2019)

\bibitem{fdsr}
Udrescu, S.M., Tegmark, M.: The feynman database for symbolic regression.
  \url{https://space.mit.edu/home/tegmark/aifeynman.html} (2020), accessed
  31.01.2020

\bibitem{verel2018sampling}
Verel, S., Daolio, F., Ochoa, G., Tomassini, M.: Sampling local optima networks
  of large combinatorial search spaces: the qap case. In: Parallel Problem
  Solving from Nature (PPSN). pp. 257--268. Springer (2018)

\bibitem{wind}
Vladislavleva, E., Friedrich, T., Neumann, F., Wagner, M.: Predicting the
  energy output of wind farms based on weather data: Important variables and
  their correlation. Renewable Energy  \textbf{50},  236 -- 243 (2013)

\bibitem{yafrani2018fitness}
Yafrani, M.E., Martins, M.S., Krari, M.E., Wagner, M., Delgado, M.R., Ahiod,
  B., L{\"u}ders, R.: A fitness landscape analysis of the travelling thief
  problem. In: Genetic and Evolutionary Computation Conference (GECCO). pp.
  277--284 (2018)

\end{thebibliography}

\end{document}